\documentclass[10pt,twocolumn,letterpaper]{article}

\usepackage[pagenumbers]{cvpr} 

\usepackage{graphicx}
\usepackage{amsmath}
\usepackage{amssymb}
\usepackage{booktabs}

\usepackage[pagebackref,breaklinks,colorlinks]{hyperref}

\usepackage[capitalize]{cleveref}
\crefname{section}{Sec.}{Secs.}
\Crefname{section}{Section}{Sections}
\Crefname{table}{Table}{Tables}
\crefname{table}{Tab.}{Tabs.}

\def\cvprPaperID{*****} 
\def\confName{CVPR}
\def\confYear{2023}

\begin{document}

\title{1st Place Solution for PVUW Challenge 2023: Video Panoptic Segmentation}

\author{Tao Zhang$^{1}${\qquad}Xingye Tian$^{2}${\qquad}Haoran Wei$^{1}${\qquad}Yu Wu$^{1}${\qquad}Shunping Ji$^{1}$\thanks{ Corresponding author.}\\
Xuebo Wang$^{2}${\qquad}Xin Tao$^{2}${\qquad}Yuan Zhang$^{2}${\qquad}Pengfei Wan$^{2}$ \vspace{3mm}\\
$^{1}$Wuhan University\qquad$^{2}$Y-tech, Kuaishou Technology 
}
\maketitle

\begin{abstract}
   Video panoptic segmentation is a challenging task that serves as the cornerstone of numerous downstream applications, including video editing and autonomous driving. We believe that the decoupling strategy proposed by DVIS enables more effective utilization of temporal information for both "thing" and "stuff" objects. In this report, we successfully validated the effectiveness of the decoupling strategy in video panoptic segmentation. Finally, our method achieved a VPQ score of 51.4 and 53.7 in the development and test phases, respectively, and ultimately ranked 1st in the VPS track of the 2nd PVUW Challenge. The code is available at \href{https://github.com/zhang-tao-whu/DVIS}{https://github.com/zhang-tao-whu/DVIS}.
\end{abstract}

\section{Introduction}
\label{sec:intro}
Video Panoptic Segmentation (VPS) is a challenging task that is extends from image panoptic segmentation \cite{vps}. VPS aims to simultaneously classify, track, segment all objects in a video, including both things and stuff. Due to its wide application in many downstream tasks such as video understanding, video editing, and autonomous driving, VPS has received increasing attention in recent years.

Video panoptic segmentation can be interpreted as a fusion of video semantic segmentation and video instance segmentation \cite{vis}. With the rapid development of deep learning, numerous exceptional works have emerged in the field of both video semantic segmentation and video instance segmentation. Recent research efforts, such as \cite{miao2021vspw, miao2023temporal, jin2021memory, he1st}, aim to enhance segmentation quality and temporal consistency in video semantic segmentation. On the other hand, video instance segmentation methods such as \cite{VITA, MinVIS, IDOL, GenVIS} are geared towards improving the quality of instance segmentation and the robustness of instance association. However, it is difficult to directly apply these methods to video panoptic segmentation due to specific designs that vary greatly. Currently, there are few works \cite{vps, vip-deeplab, video-knet} focusing on video panoptic segmentation, and most of them are simply extended from image panoptic segmentation methods \cite{upsnet, panoptic-deeplab, knet}.

Recently, DVIS \cite{DVIS} has decoupled the task of video instance segmentation into three independent sub-tasks: image instance segmentation, tracking/alignment, and refinement. In addition, DVIS has designed a referring tracker and a temporal refiner to achieve stable tracking and optimal utilization of temporal information, which has demonstrated great advantages in the VIS field. A natural question is whether DVIS can be used for video panoptic segmentation or universal segmentation. In this report, we found that DVIS exhibits equally excellent performance on stuff and thing objects. Stuff objects have no difference with thing objects, and their deformation in the temporal dimension is lighter and motion trajectory is simpler. Therefore, DVIS has achieved state-of-the-art performance in video panoptic segmentation without any modification.

Thanks to the superior performance of DVIS, we achieved first place in the VPS track of the 2nd PVUW challenge in CVPR 2023. DVIS achieved 51.5 VPQ during development and 53.7 VPQ during testing, without using any additional training data (including the validation set of VIPSeg).
Thanks to the superior performance of DVIS, we were able to achieve first place in the VPS track of the 2nd PVUW challenge at CVPR 2023. DVIS achieved 51.5 VPQ and 53.7 VPQ in the development and test phases without using any additional training data, including the validation set of VIPSeg.

\begin{figure}[t]
  \centering
   \includegraphics[width=1\linewidth]{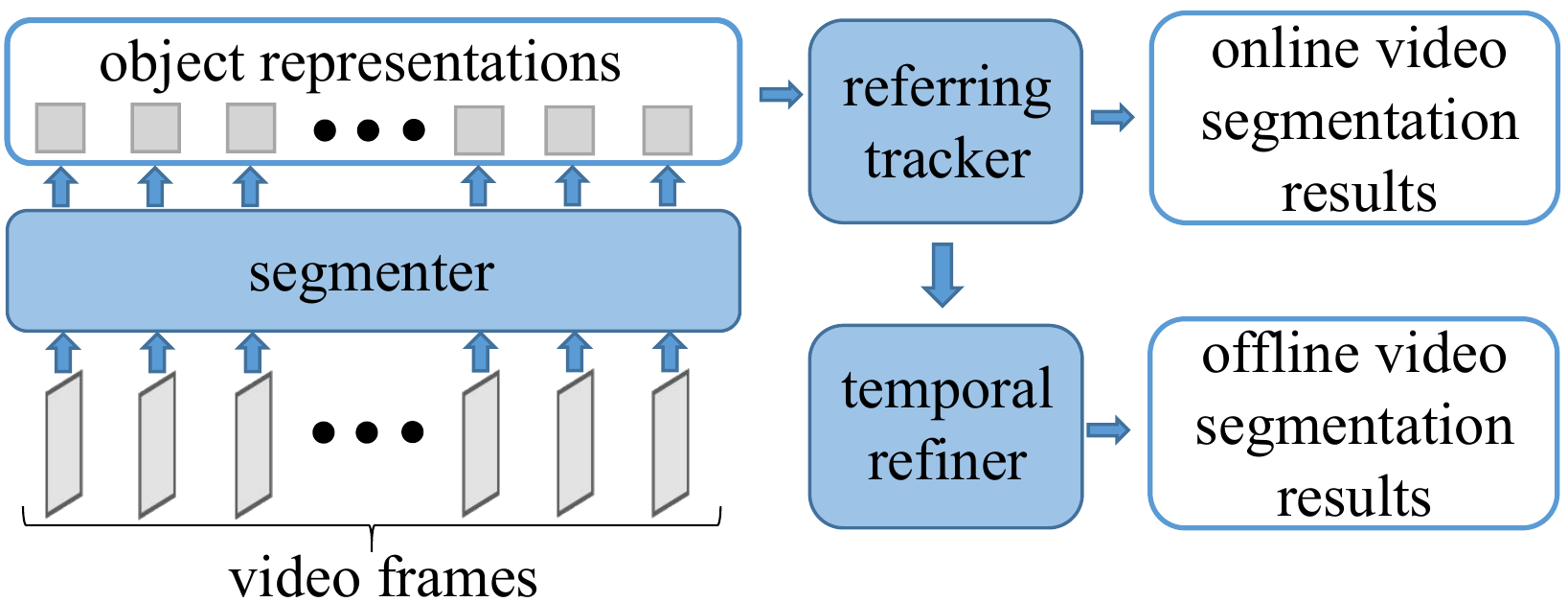}
   \caption{\textbf{Our framework overview.} It consists of three independent components: a segmenter, a referring tracker, and a temporal refiner.
   }
   \label{fig:framework}
\end{figure}

\section{Method}
DVIS propose a novel decoupled framework for Video Segmentation that consists of three independent components: a segmenter, a referring tracker, and a temporal refiner, as illustrated in \cref{fig:framework}.
The segmenter in our framework, as illustrated in Section \ref{sec: segmenter}, is introduced. The referring tracker is described in Section \ref{sec: reftracker}, while the temporal refiner is presented in Section \ref{sec:tmpref}.
\begin{figure*}[t]
  \centering
   \includegraphics[width=1\linewidth]{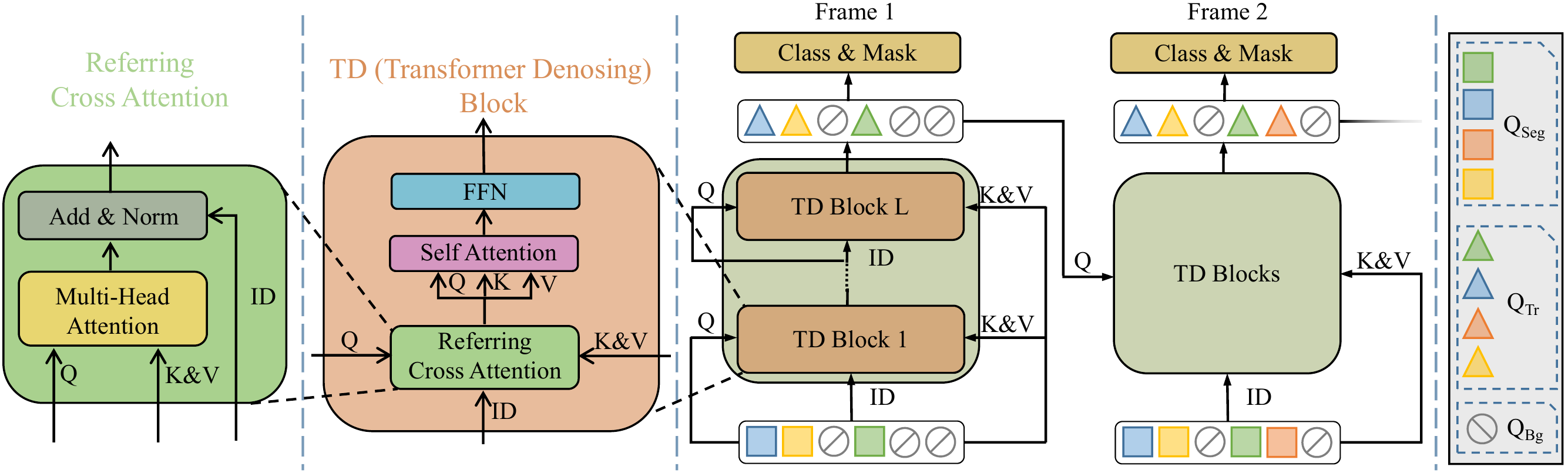}
   \caption{\textbf{The framework of the referring tracker.} The object representations output by the segmenter $(Q_{seg})$and referring tracker
$(Q_{Tr})$ are represented by squares and triangles, respectively. Objects with the same ID are assigned the same color.
   }
   \label{fig:retrack}
\end{figure*}

\begin{figure}[t]
  \centering
   \includegraphics[width=1\linewidth]{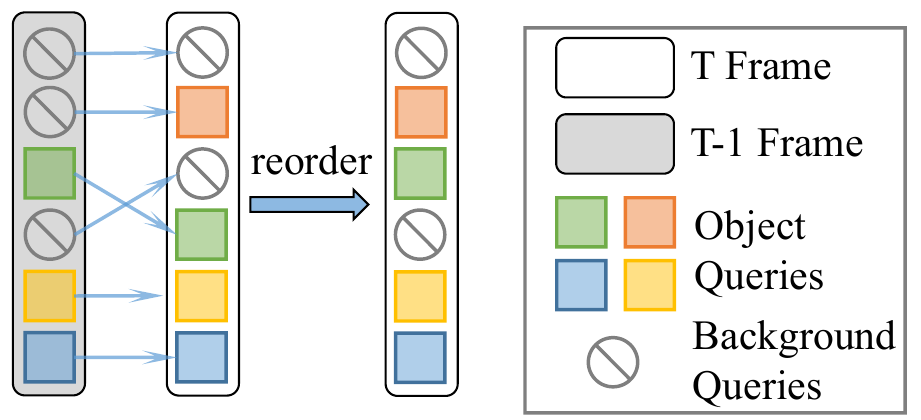}
   \caption{\textbf{The pre-matching process.}}
   \label{fig:mathcing}\vspace{-1mm}
\end{figure}
\par
\subsection{Segmenter \label{sec: segmenter}}
DVIS employ Mask2Former \cite{mask2former} as the segmenter in our framework. Mask2Former is a universal image segmentation architecture that surpasses specialized architectures in various segmentation tasks while maintaining ease of training for each specific task. It is constructed upon a straightforward meta-architecture comprising a backbone, a pixel decoder, and a transformer decoder. Notable enhancements encompass masked attention within the Transformer decoder, which confines attention to localized features centered around predicted segments, as well as the integration of multi-scale high-resolution features that facilitate precise segmentation of small objects and regions.

\subsection{Referring Tracker \label{sec: reftracker}}
The referring tracker employs the modeling paradigm of referring denoising to tackle the inter-frame correlation task. Its main objective is to leverage denoising operations to optimize the initial values and generate more accurate tracking results.
\par
The referring tracker comprises a sequence of $L$ transformer denoising (TD) blocks, each composed of a reference cross-attention (RCA), a standard self-attention, and a feed-forward network (FFN).~\cref{fig:retrack} illustrates the architecture of the referring tracker.~It takes object queries $ \left\{ Q^i_{seg}| i \in [1,T] \right\}$ generated by the segmenter as input and produces object queries $ \left\{ Q^i_{Tr}| i \in [1,T] \right\}$ for the current frame that corresponds to objects in the previous frame. In this context, T represents the length of the video.
\par
Firstly, as shown in \cref{fig:mathcing}, the Hungarian matching algorithm \cite{hungarian} is utilized to match the $Q_{seg}$ of adjacent frames, as is done in \cite{MinVIS}.
\begin{equation}
  \left\{
    \begin{aligned}
    \widetilde{Q}_{seg}^i & = Hungarian(\widetilde{Q}_{seg}^{i-1},Q_{seg}^i), i \in [2,T]  \\
    \widetilde{Q}_{seg}^i & = Q_{seg}^i, i=1 \\
    \end{aligned}
    \right.
\end{equation}
Where $\widetilde{Q}_{seg}$ represents the matched object query generated by the segmenter. $\widetilde{Q}_{seg}$ can be regarded as a tracking result with noise and serves as the initial query for the reference tracker. In order to remove noise from the initial query $\widetilde{Q}_{seg}^i$ of the current frame, the reference tracker leverages the denoised object query $Q_{Tr}^{i-1}$ from the previous frame as a reference.
\par
Next, $\widetilde{Q}_{seg}^i$ is fed into the TD block, where the crucial denoising process is performed using RCA, resulting in the output $Q^i_{Tr}$.
RCA serves as the core component of the referring tracker, effectively leveraging the similarity between object representations of adjacent frames while mitigating potential confusion caused by their similarity.~Given that the appearance of the same object in adjacent frames tends to be similar while its position, shape, and size may vary, initializing the object representation of the current frame with the representation from the previous frame \cite{GenVIS} introduces ambiguity. To address this issue, RCA incorporates an identity (ID) mechanism, which effectively exploits the similarity between the query (Q) and key (K) to generate accurate outputs.~\cref{fig:retrack} illustrates the inspiration behind RCA and its slight modifications in comparison to standard cross-attention:
\begin{equation}
RCA(ID,Q,K,V)=ID+MHA(Q,K,V)
\end{equation}
MHA refers to Multi-Head Attention \cite{transformer}, while ID, Q, K, and V denote identification, query, key, and value, respectively.
\par
Finally, the denoised object query $Q^i_{Tr}$ is employed as an input for both the class head and mask head. The class head generates the category output, while the mask head produces the mask coefficient output.
\subsection{Temporal Refiner \label{sec:tmpref}}
The limitations of previous offline video segmentation methods primarily stem from the limited utilization of temporal information by tightly coupled networks. Additionally, the current online methods lack a refinement step. To overcome these challenges, we propose an independent temporal refiner. This module efficiently leverages the temporal information across the entire video and refines the output generated by the referring tracker.
\par
\begin{figure}[t]
  \centering
   \includegraphics[height=85mm,width=85mm]{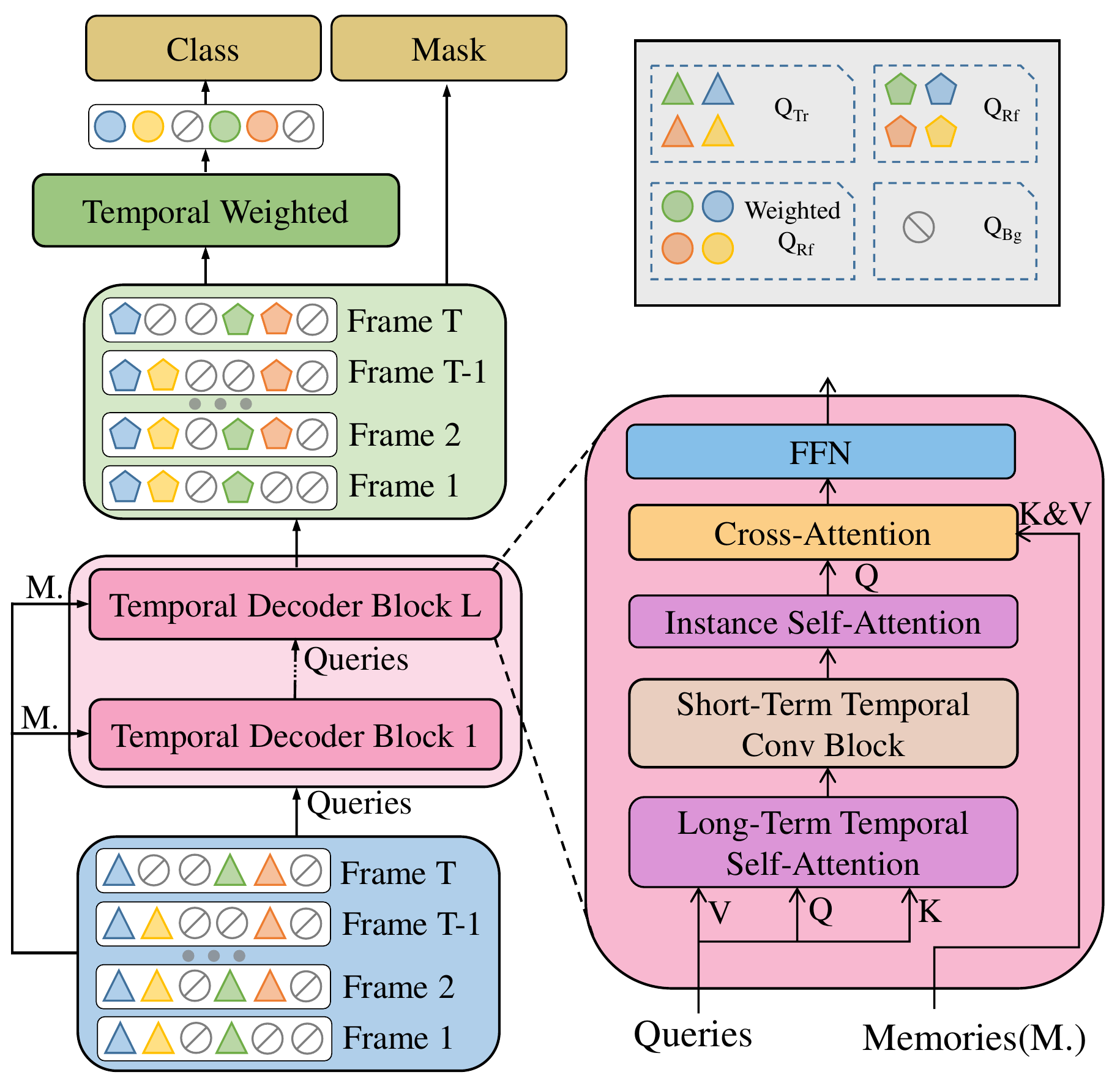}
   \caption{
   \textbf{The framework of the temporal refiner.} 
   Instance representations for each frame ($Q_{Rf}$) are denoted by pentagons, while the object representations for the entire video ($\widetilde{Q}_{Rf}$ ) are denoted by circles. 
   Different colors indicate different object IDs, and gray represents the background. 
   “$L_{\times}$” means the block repeated $L$ times.
   }
   \label{fig:temref} \vspace{-3mm}
\end{figure}

\cref{fig:temref} illustrates the architecture of the temporal refiner, which plays a crucial role in enhancing the temporal information utilized by the model. The temporal refiner takes the object query $Q_{Tr}$ generated by the reference tracker as input and produces the refined object query $Q_{Rf}$ by aggregating temporal information from the entire video. The temporal refiner is composed of L temporal decoder blocks connected in a cascaded operation. Each temporal decoder block consists of two key components: a short-term temporal convolutional block and a long-term temporal attention block. The short-term temporal convolutional block leverages motion information, while the long-term temporal attention block integrates information from the entire video. These components employ 1D convolutions and standard self-attention, respectively, operating on the temporal dimension.
\par
Finally, the mask head generates mask coefficients for each object in every frame, utilizing the refined object query $Q_{Rf}$.~Additionally, the class head utilizes the temporal weights of $Q_{Rf}$ to predict the class and score of each object across the entire video.~The temporal weighting process can be defined as follows:
\begin{equation} \vspace{-2mm}
\hat{Q}_{Rf}=\sum\limits_{t=1}^{n} 
Softmax(Linear(Q^t_{Rf}))Q^t_{Rf}
\vspace{-1mm}
\end{equation}
where $\hat{Q}_{Rf}$ is the temporal weighting of $Q_{Rf}$.
\subsection{Loss \label{sec:loss}}
Specifically,~our training focuses on the referring tracker and temporal refiner components, while we freeze the segmenter.
Given that the referring tracker operates on a frame-by-frame basis, its supervision relies on a loss function tailored to this approach. Specifically, object labels and predictions $\hat{y}_{Tr}$ are matched only on the frame where the object initially appears. To expedite convergence during the early stages of training, predictions from the frozen segmenter $\hat{y}_{seg}$ are utilized for matching instead of the referring tracker's predictions.
\begin{equation}
  \left\{
    \begin{aligned}
    \hat{\sigma} & =  \mathop{argmin}\limits_{\sigma} \sum\limits_{i=1}^N \mathcal L_{match}(y_i^{f(i)},\hat{y}_{\sigma(i)}^{f(i)}) \\
    \hat{y} & =\hat{y}_{Tr} \ if \ Iter \ge \frac{Max_Iter}{2} \ else \ \hat{y}_{seg}
    \end{aligned}
    \right.
\end{equation}
where $f(i)$ represents the frame in which the $i_{th}$ instance first appears. $\mathcal L(y_i^{f(i)},\hat{y}_{\sigma(i)}^{f(i)})$ is a pair-wise matching cost, as used in \cite{mask2formervis}, between the ground truth $y$ and the prediction $\hat{y}$ having index $\sigma(i)$ on the $f(i)$ frame.
The loss function $\mathcal L$ is exactly the same as that in \cite{mask2formervis}.
\begin{equation}
\mathcal L_{Tr}=\sum\limits_{t=1}^T\sum\limits_{i=1}^N\mathcal L(y_i^{t},\hat{y}_{\hat{\sigma}(i)}^{t})
\end{equation}
The temporal refiner is supervised during training using the same matching cost and loss functions as \cite{mask2formervis}. 
The segmenter and the referring tracker are frozen during training. Consequently, the referring tracker's prediction results are employed for matching during the initial training phase, guiding the network toward accelerated convergence.
\begin{equation}
  \left\{
    \begin{aligned}
    \hat{\sigma} & =  \mathop{argmin}\limits_{\sigma} \sum\limits_{i=1}^N \mathcal L_{match}(y_i,\hat{y}_{\sigma(i)}) \\
    \hat{y} & =\hat{y}_{Rf} \ if \ Iter \ge \frac{Max_Iter}{2} \ else \ \hat{y}_{Tr}
    \end{aligned}
    \right.
\end{equation}
where $\hat{y}_{Rf}$ is the prediction result of the temporal refiner.
The loss function of temporal refiner is:
\begin{equation}
\mathcal L_{Rf}=\sum\limits_{i=1}^N\mathcal L(y_i,\hat{y}_{\hat{\sigma}(i)})
\end{equation}
\section{Experiment}
\subsection{Implementation Details}
In our approach, we employ Swin Large \cite{swin} as the backbone and Mask2Former as the segmenter for video segmentation. The segmenter, referring tracker, and temporal refiner are trained separately. We fine-tune the segmenter with COCO \cite{coco} pre-trained weights using image-level annotations from the training set of VIPSeg \cite{vipseg}. During referring tracker training, we freeze the segmenter and use a continuous 5-frame clip from the video as input. In the case of training the temporal refiner, we freeze both the segmenter and the referring tracker and use a continuous 21-frame clip as input. Training is carried out on the training set of VIPSeg (excluding additional data such as validation set) for 20k iterations with a batch size of 8, and the learning rate is decayed by 0.1 at 14k iterations. Multi-scale training is used to randomly scale the short side of input video clips from 480 to 800 during training. Additionally, for training the refiner, we employ a random cropping strategy involving tiles of size 608$\times$608 from input video clips. Our training is executed on 8 NVIDIA 4090 GPUs, with fine-tuning of the segment requiring 18GB of GPU memory and taking approximately 3 hours. Furthermore, the training process for the referring tracker requires 7GB of GPU memory and takes around 7 hours. Finally, the temporal refiner necessitates 8GB of GPU memory and takes approximately 15 hours to train.
\begin{table}[t]
\setlength{\tabcolsep}{1.4mm}
\centering
\begin{tabular}{l|cccccc}
	Method & VPQ & VPQ1 & VPQ2 & VPQ4 & VPQ6 & STQ   \\
	\hline
	Ours & \textbf{51.4} & \textbf{52.1} & \textbf{51.5} &	\textbf{51.2} & \textbf{51.1} & 0.47 \\
    yknykn & 49.6 & 50.7 & 49.7 & 49.1 & 48.8 & 0.49 \\
    francis\_fan & 49.0 & 50.1 &	49.1 & 48.5 & 48.2 & 0.48 \\
    korpusose & 46.2 & 49.7 & 47.0	& 44.9 & 43.1 & 0.46 \\ 
    DaiSir & 45.4 &	46.5 & 45.8 & 45.0 & 44.5 & \textbf{0.49} \\
    llredahhh & 45.0 & 49.2 & 45.4 & 43.4 & 42.0 & 0.42 \\
    dlsrbgg33 & 45.0 & 49.2 & 45.8 & 43.4 & 41.6 & 0.47 \\
    yyyds & 43.2 & 44.3 & 43.6 & 42.8 & 42.1 & 0.47 \\
\hline
 \end{tabular}
\caption{\textbf{Leaderboard during the development phase.}}
 \label{tab:develop}
\end{table}

\begin{table}[t]
\setlength{\tabcolsep}{1.4mm}
\centering
\begin{tabular}{l|cccccc}
	Method & VPQ & VPQ1 & VPQ2 & VPQ4 & VPQ6 & STQ   \\
	\hline
	Ours & \textbf{53.7} & 54.7 & \textbf{54.1} & \textbf{53.3} & \textbf{52.8} & 0.51 \\
    yknykn & 52.9 & 54.4 & 53.1 & 52.3 & 51.8 & 0.52 \\
    llredahhh & 50.1 & \textbf{55.3} & 50.7 & 48.1 & 46.4 & 0.46 \\
    yyyds & 50.0 & 51.6 & 50.6 & 49.4 & 48.5 & 0.52 \\
    SUtech & 49.9 & 51.6 & 50.6 & 49.2 & 48.1 & \textbf{0.52} \\
    korpusose & 48.6 & 52.8 & 49.8 & 46.9 & 44.8 & 0.48 \\
    DaiSir & 47.1 & 48.7 & 47.7 & 46.5 & 45.6 &	0.52 \\
    AA250250 & 46.0 & 47.6 & 46.6 &	45.4 & 44.5 & 0.51 \\
    Fancy\_z & 45.9 & 47.8 & 46.6 & 45.2 & 44.0 & 0.51 \\
\hline
 \end{tabular}
\caption{\textbf{Leaderboard during the test phase.}}
 \label{tab:test}
\end{table}
\subsection{Comparison with Other Methods}
In the second PVUW Challenge, we ranked first in both the development and test phases. The leaderboards for the development and test phases are displayed in tables 1 and 2, respectively. Our method achieved a VPQ of 51.4 in the development phase and 53.7 in the test phase, surpassing all other methods. Additionally, our method has significant advantages in tracking stability. Compared to VPQ1, our method's VPQ6 only decreased by 1.0 and 1.9 in the development and test phases, respectively. In contrast, the second and third-place methods in the development phase showed a decrease of 1.9, and the second and third-place methods in the test phase exhibited a decrease of 2.6 and 8.9, respectively.
\subsection{Ablation Study}
\begin{table}[t]
\setlength{\tabcolsep}{1.4mm}
\centering
\begin{tabular}{l|cccccc}
	Method & VPQ & VPQ1 & VPQ2 & VPQ4 & VPQ6\\
    \hline
    Baseline & 52.6 & 53.6 & 52.9 & 52.2 & 51.8\\
	+Multi Scale & 53.7 & 54.7 & 54.1 & 53.3 & 52.8\\
\hline
 \end{tabular}
\caption{\textbf{Ablation study of our applied modules.}}
 \label{tab:ablation} \vspace{-2mm}
\end{table}
During the test phase, we utilized a multi-scale testing augmentation approach. The input video was scaled to resolutions of 720p and 800p, and the resulting prediction results were combined to form the final prediction outcome. This resulted in a 1.1 VPQ performance improvement.

\section{Conclusion}
We introduced DVIS to the VPS field and verified that the decoupling strategy proposed by DVIS significantly improved the performance for both thing and stuff objects. As a result, we won the championship in the VPS track of the 2nd PVUW Challenge, scoring 51.4 VPQ and 53.7 VPQ in the development and test phases, respectively.
{\small
\bibliographystyle{ieee_fullname}
\bibliography{egbib}
}

\end{document}